\documentclass[10pt,twocolumn,letterpaper]{article}

\usepackage{iccv}
\usepackage{times}
\usepackage{epsfig}
\usepackage{graphicx}
\usepackage{amsmath}
\usepackage{amssymb}

\usepackage{booktabs}
\usepackage{multirow}
\usepackage{multicol}
\usepackage{colortbl}

\def\eg{\emph{e.g.}}
\def\ie{\emph{i.e.}}
\def\etc{\emph{etc}}
\def\vs{\emph{vs}}
\def\ourmethod{SgMg}
\usepackage[pagebackref=true,breaklinks=true,letterpaper=true,colorlinks,bookmarks=false]{hyperref}

\iccvfinalcopy 

\def\iccvPaperID{8733} 

\ificcvfinal\fi

\begin{document}

\title{Spectrum-guided Multi-granularity Referring Video Object Segmentation}

\author{Bo Miao$^{1}$,~
Mohammed Bennamoun$^{1}$,~
Yongsheng Gao$^{2}$,~
Ajmal Mian$^{1}$
\\[0.1cm]
${^1}$The University of Western Australia~~~
${^2}$Griffith University
\vspace{4mm}
\\
}

\maketitle
\ificcvfinal\thispagestyle{empty}\fi

\begin{abstract}
Current referring video object segmentation (R-VOS) techniques extract conditional kernels from encoded (low-resolution) vision-language features to segment the decoded high-resolution features. We discovered that this causes significant feature drift, which the segmentation kernels struggle to perceive during the forward computation. This negatively affects the ability of segmentation kernels. 
To address the drift problem, we propose a Spectrum-guided Multi-granularity (\ourmethod{}) approach, which performs direct segmentation on the encoded features and employs visual details to further optimize the masks.
In addition, we propose Spectrum-guided Cross-modal Fusion (SCF) to perform intra-frame global interactions in the spectral domain for effective multimodal representation.
Finally, we extend \ourmethod{} to perform multi-object R-VOS, a new paradigm that enables simultaneous segmentation of multiple referred objects in a video. This not only makes R-VOS faster, but also more practical. Extensive experiments show that \ourmethod{} achieves state-of-the-art performance on four video benchmark datasets, outperforming the nearest competitor by 2.8\% points on Ref-YouTube-VOS. Our extended \ourmethod{} enables multi-object R-VOS, runs about 3$\times$ faster while maintaining satisfactory performance. Code is available at \href{https://github.com/bo-miao/SgMg}{https://github.com/bo-miao/SgMg}.
\end{abstract}

\section{Introduction}
\label{sec:intro}
Referring video object segmentation (R-VOS) aims at segmenting objects in a video, referred to by linguistic descriptions. 
R-VOS is an emerging task for multimodal reasoning and promotes a wide range of applications, including language-guided video editing and human-machine interaction.
Different from conventional semi-supervised video object segmentation~\cite{STM,STCN,RAVOS}, where the mask annotation for the first frame is provided for reference, R-VOS is more challenging due to the need for cross-modal understanding between vision and free-form language expressions.

\begin{figure}[t]
\centering
\includegraphics[width=\columnwidth]{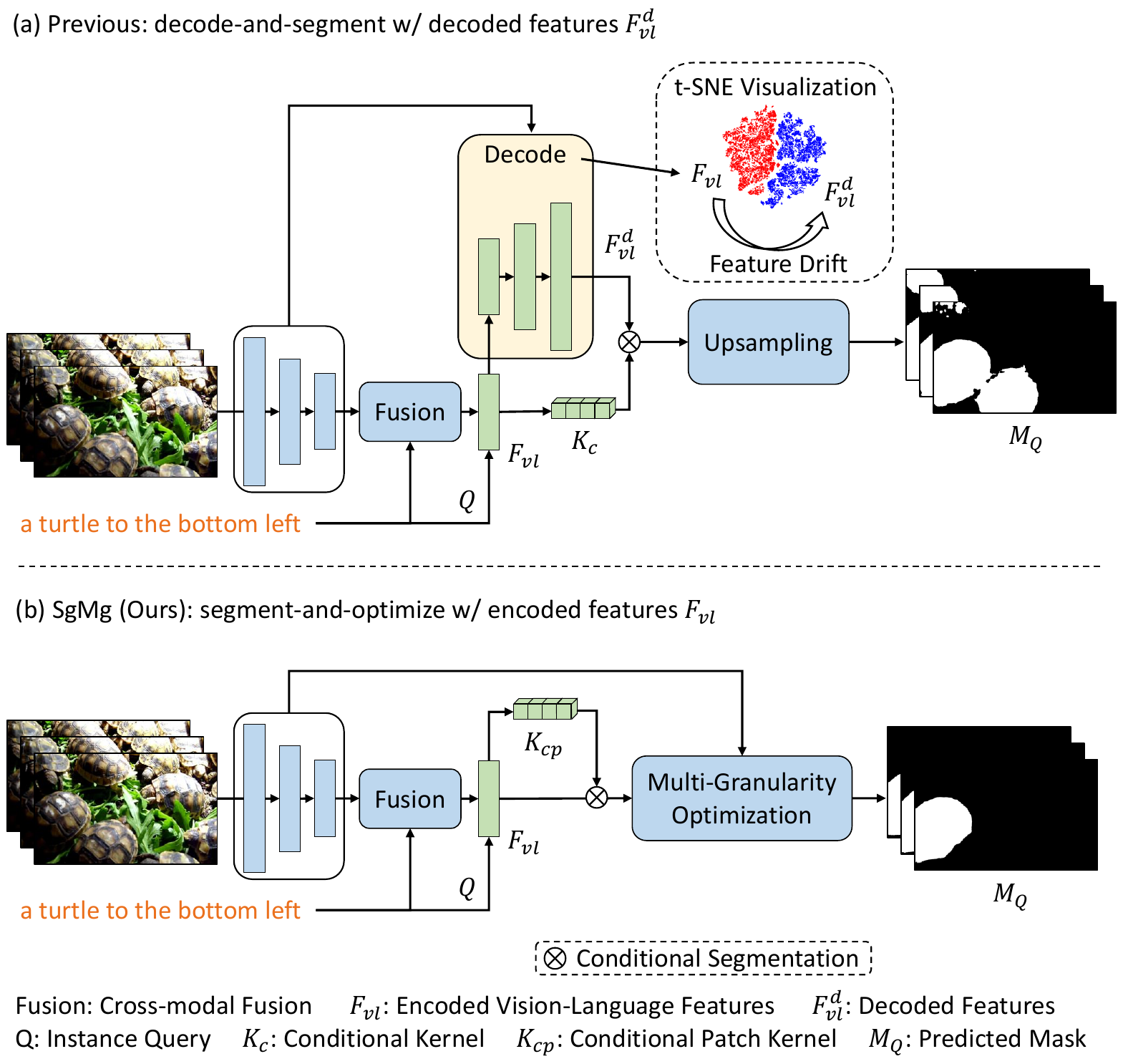}
\caption{(a) Previous methods~\cite{MTTR,ReferFormer} apply segmentation kernels $\mathcal{K}_{c}$~\cite{ConditionConv}, extracted from encoded features $\mathcal{F}_{vl}$, to segment the decoded high-resolution features $\mathcal{F}^{d}_{vl}$. 
(b) We use segmentation kernels $\mathcal{K}_{cp}$, extracted from encoded features $\mathcal{F}_{vl}$, to segment the encoded features $\mathcal{F}_{vl}$ directly, and propose multi-granularity optimization to recover visual details and produce fine-grained masks.}
\label{fig:paradigm}
\end{figure}

Early R-VOS techniques~\cite{RefVOS,RefDAVIS,HINet} perform feature encoding, cross-modal interaction, and language grounding using convolutional neural networks (CNNs). However, the limited ability of CNNs to capture long-range dependencies and handle free-form features constrains the model performance. 
With the advancement of attention mechanisms~\cite{Attention,OTS,ScenceAttn}, recent methods achieved significant improvement on R-VOS using cross-attention~\cite{URVOS,YOFO,R2VOS} for multimodal understanding and transformers~\cite{MANet,MLRL} for spatio-temporal representation. 
Based on transformers, conditional kernel~\cite{ConditionConv} is then introduced to separate foreground from semantic features given its high adaptability to different instances~\cite{MTTR,ReferFormer}. 
As illustrated in Fig.~\ref{fig:paradigm}(a), these methods attend to encoded vision-language features $\mathcal{F}_{vl}$ using instance queries $Q$ to predict conditional kernels $\mathcal{K}_{c}$, and employ $\mathcal{K}_{c}$ as the segmentation head to segment decoded features $\mathcal{F}_{vl}^{d}$. Despite the promising performance, this paradigm still has some limitations. 
\emph{Firstly}, as shown in the t-SNE~\cite{tsne} visualization in Fig.~\ref{fig:paradigm}(a), although the nonlinear decoding process introduces visual details, this is accompanied by a significant feature drift, which increases the difficulty of segmentation since $\mathcal{K}_{c}$ is predicted before feature decoding.
\emph{Secondly}, bilinear upsampling of the predicted masks $\mathcal{M}_{Q}$ to increase resolution impedes the segmentation performance.
\emph{Thirdly}, these methods only support single expression-based segmentation, making R-VOS inefficient when multiple referred objects exist in a video.

In this work, we propose a Spectrum-guided Multi-granularity (\ourmethod{}) approach that follows a   segment-and-optimize pipeline to address the above problems. 
As depicted in Fig.~\ref{fig:paradigm}(b), \ourmethod{} introduces Conditional Patch Kernel (CPK) $\mathcal{K}_{cp}$ to directly segment its fully perceived encoded features $\mathcal{F}_{vl}$, avoiding the feature drift and its adverse effects.
The segmentation is then refined using our proposed Multi-granularity Segmentation Optimizer (MSO), which employs low-level visual details to produce full-resolution masks.
Within the \ourmethod{} framework, we further develop Spectrum-guided Cross-modal Fusion (SCF) that performs intra-frame global interactions in the spectral domain to facilitate multimodal understanding.
Finally, we introduce a new paradigm called multi-object R-VOS to simultaneously segment multiple referred objects in a video. To achieve this, we extend \ourmethod{} by devising multi-instance fusion and decoupling. Our main contributions are summarized as follows: 

\begin{itemize}
\item[$\bullet$] We explain how existing R-VOS methods suffer from the feature drift problem. To address this problem, we propose \ourmethod{} that follows a segment-and-optimize pipeline and achieves top-ranked overall performance on multiple benchmark datasets.
\item[$\bullet$] We propose Spectrum-guided Cross-modal Fusion to encourage intra-frame global interactions in the spectral domain.
\item[$\bullet$] We extend \ourmethod{} to perform multi-object R-VOS, a new paradigm that enables simultaneous segmentation of multiple referred objects in a video. Our multi-object variant is more practical and runs $3\times$ faster.
\end{itemize} 

We conduct extensive experiments on multiple benchmark datasets, including Ref-YouTube-VOS~\cite{URVOS}, Ref-DAVIS17~\cite{RefDAVIS}, A2D-Sentences~\cite{GavrilyukA2D}, and JHMDB-Sentences~\cite{JHMDB}, and achieve state-of-the-art performance on all four.
On the largest validation set Ref-YouTube-VOS, \ourmethod{} achieves 65.7 $\mathcal{J}$\&$\mathcal{F}$ which is 2.8\% points higher than that of the closest competitor ReferFormer~\cite{ReferFormer}. 
On the A2D-Sentences, \ourmethod{} achieves 58.5 mAP which is 3.5\% points higher than that of ReferFormer.

\begin{figure*}[t!]
\centering
\includegraphics[width=2\columnwidth]{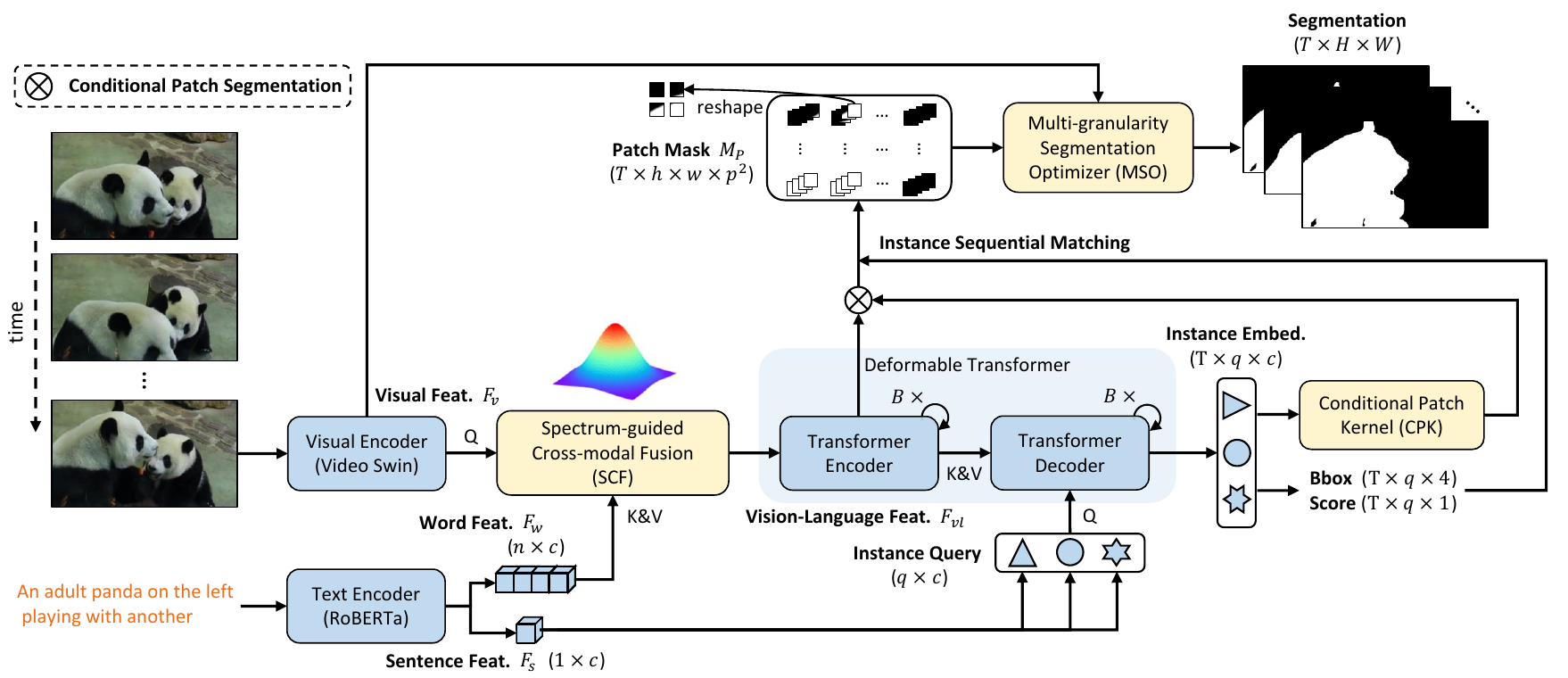}
\vspace{1mm}
\caption{The overall framework of \ourmethod{}. Taking a video sequence $\mathcal{V} = \{I_{i}\}_{i=1}^T$ and a language expression $\mathcal{L} = \{S_{i}\}_{i=1}^N$ as input, \ourmethod{} predicts the masks of referred object $\mathcal{O}_{\mathcal{L}}$ in each frame. 
SCF projects visual features $\mathcal{F}_{v}$ to vision-language features $\mathcal{F}_{vl}$, instance-aware CPK predicts patch masks by segmenting encoded $\mathcal{F}_{vl}$, and MSO optimizes patch masks to get fine-grained results.}\label{fig:framework}
\end{figure*}

\section{Related Works}

\noindent\textbf{Video Object Segmentation} techniques fall into two categories: unsupervised and semi-supervised. 
Unsupervised approaches segment the most salient instances in each video without user interactions~\cite{UVOS1,UVOS2}. They often employ two-stream networks to fuse motion and appearance cues for segmentation. 
Semi-supervised approaches track the given first frame object mask by performing online learning~\cite{OSVOS} or spatial-temporal association~\cite{STM,STCN,AOT,MAMP,FEELVOS}. Unlike conventional semi-supervised video object segmentation, R-VOS takes a free-form linguistic expression as guidance to detect and segment referred objects in videos.

\noindent\textbf{Referring Video Object Segmentation.} 
R-VOS methods mainly use deep neural networks with vision-and-language interaction to empower visual features with corresponding linguistic information for pixel-level segmentation. 
For example,~\cite{URVOS} employs a unified R-VOS framework that performs iterative segmentation guided by both language and temporal features. \cite{CMPC,Clawcranenet} adopt progressive segmentation by perceiving potential objects and discriminating the best match. \cite{MMTBVS} fuses visual and motion features for segmentation under the guidance of linguistic cues. \cite{CITD} models object relations to form tracklets and performs tracklet-language grounding. 
To enhance multi-modal interactions, \cite{HINet, DITE, PMINet,LBDT} perform hierarchical vision-language fusion on multiple feature layers. 

Despite their promising performance, the complex multi-stage pipelines and use of multiple networks make R-VOS burdensome. 
To address these problems, MTTR~\cite{MTTR} proposes an end-to-end transformer-based network with conditional kernels~\cite{ConditionConv} to segment target objects.
ReferFormer~\cite{ReferFormer} further introduces language-guided instance queries to predict instance-aware conditional kernels and an auxiliary detection task to aid localization. 
These methods follow a decode-and-segment pipeline, which adopts conditional kernels to segment decoded high-resolution features to achieve promising performance. 
However, the nonlinear decoding process leads to significant feature drift that negatively affects the conditional kernels.
In contrast to previous works, our approach follows a segment-and-optimize pipeline to avoid the adverse drift effects and to predict full-resolution masks in an efficient manner.

\noindent\textbf{Vision and Language Representation Learning} aims to learn vision-language semantics and alignment for multimodal reasoning tasks. It has achieved significant success in various tasks~\cite{12-in-1,Vinvl,TTA}, including video question answering~\cite{VL-VQA}, video captioning~\cite{VC}, video-text retrieval~\cite{VSE,CenterCLIP}, zero-shot classification~\cite{CLIP}, referring image/video segmentation~\cite{URVOS}, \etc. 
Some approaches~\cite{CLIP,AlignFuse-VQA} rely on contrastive pre-training using large-scale datasets to project different modalities into unified embedding space.
Others~\cite{vilbert,EFNet} develop cross-modal interaction layers for multimodal feature fusion and understanding. 
Recent deep learning methods in spectral domain~\cite{MLSpectrum,DSM,FFC,GSFM} have raised widespread awareness because of their ability to perform global interactions. 
We take inspiration from these spectral-based methods and employ spectrum guidance in the field of vision-language representation to encourage multimodal global interactions.

\section{\ourmethod{}: Spectrum-guided Multi-granularity Referring Video Object Segmentation}

Given a video sequence $\mathcal{V} = \{I_{i}\}_{i=1}^T$ with $T$ frames and a language query $\mathcal{L} = \{S_{i}\}_{i=1}^N$ with $N$ words. The goal of R-VOS is to segment the referred object $\mathcal{O}_{\mathcal{L}}$ in $\mathcal{V}$ at pixel-level. To this end, we introduce a new approach termed \ourmethod{}. Different from previous R-VOS methods~\cite{MTTR,ReferFormer}, 
our approach follows a segment-and-optimize pipeline.

An overview of \ourmethod{} is shown in Fig.~\ref{fig:framework}. 
Video Swin Transformer~\cite{VideoSwin} is adopted to extract visual feature $\mathcal{F}_{v}$ and RoBERTa~\cite{Roberta} is adopted to extract sentence $\mathcal{F}_{s}$ and word $\mathcal{F}_{w}$ features. The channel dimension of all features is projected to 256. 
Spectrum-guided Cross-modal Fusion (SCF) cross attends $\mathcal{F}_{v}$ with $\mathcal{F}_{w}$ to compute vision-language features $\mathcal{F}_{vl}$. 
Deformable Transformer~\cite{DeformableTransformer} encoder is used to encode $\mathcal{F}_{vl}$ and the decoder associates instance queries created based on $\mathcal{F}_{s}$ to predict instance embeddings and the corresponding Conditional Patch Kernels (CPKs). 
Finally, the CPKs are employed to segment $\mathcal{F}_{vl}$ and predict patch masks that are further optimized with visual details through Multi-granularity Segmentation Optimizer (MSO).
The choice of the encoder and transformer follows previous works to avoid distractions~\cite{MTTR,ReferFormer}.

\subsection{Feature Drift Analysis} 
\label{sec:driftanalysis}
Existing R-VOS methods~\cite{ReferFormer,MTTR} follow a decode-and-segment pipeline where conditional kernels $\mathcal{K}_{c}$~\cite{ConditionConv} are extracted from encoded features $\mathcal{F}_{vl}$ and used to segment the decoded features $\mathcal{F}_{vl}^{d}$.
However, the decoding process leads to feature drift, which is evident in the t-SNE visualization depicted in Fig.~\ref{fig:paradigm}(a). This drift is difficult for the kernels $\mathcal{K}_{c}$ to perceive during the forward computation since $\mathcal{K}_{c}$ is predicted before the feature decoding. Therefore, we argue that \emph{even though the feature decoding enhances visual details, it also causes the drift problem that negatively affects the segmentation kernels}. This makes the existing decode-and-segment pipeline sub-optimal. 

To overcome the adverse effects of feature drift while recovering visual details, we present \ourmethod{}, a novel approach that follows a \emph{segment-and-optimize} pipeline.
In a nutshell, \ourmethod{} performs Spectrum-guided Cross-modal Fusion to compute $\mathcal{F}_{vl}$, leverages Conditional Patch Kernels to segment encoded features $\mathcal{F}_{vl}$ to avoid the drift effects, and recovers visual details with Multi-granularity Segmentation Optimizer to generate fine-grained masks.

\subsection{Spectrum-guided Cross-modal Fusion}

The two-dimensional discrete Fourier transform converts spatial data into the spectral domain. Based on the spectral convolution theorem~\cite{FFTTheory}, point-wise update of signals in the spectral domain globally affects all inputs in the spatial domain, which gives the insight to design spectrum-based modules so as to efficiently facilitate global interactions, which is critical for multimodal understanding.
In addition, Low-frequency components in the spectral domain usually correspond to the general semantic information according to previous theoretical studies \cite{DNNFreq2,DNNFreq,DNNFreq3}.

\begin{figure}[t]
\centering
\includegraphics[width=1\columnwidth]{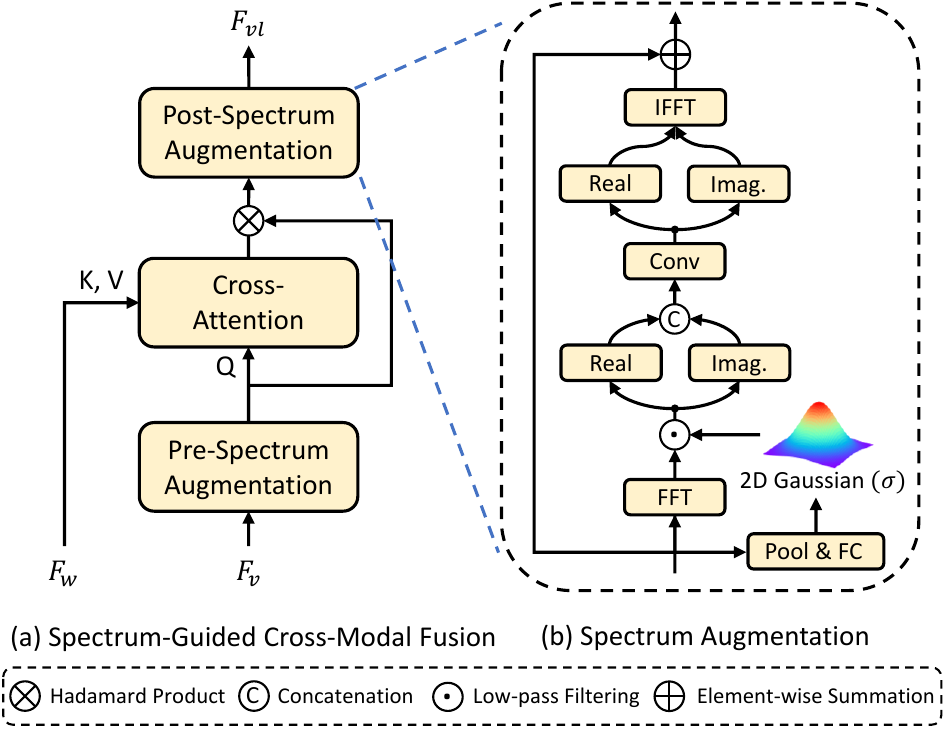}
\caption{Spectrum-guided Cross-modal Fusion. Imag.: Imaginary. Pre-spectrum augmentation and post-spectrum augmentation share an identical structure.}
\label{fig:fusion}
\end{figure}

Inspired by the above observations, we conjecture that low-frequency components can benefit higher dimensional semantic features and propose Spectrum-guided Cross-modal Fusion (SCF). 
As shown in Fig.~\ref{fig:fusion}, SCF performs pre-spectrum augmentation to enhance visual features before cross-modal fusion and post-spectrum augmentation to facilitate global vision-language interactions after the fusion process. 
Let $\mathcal{F} \in{\mathbb{R} ^{C \times H \times W}}$ denotes the input features, the spectrum augmentation (SA) is computed as:
\begin{small}
\begin{equation}
{\rm SA}(\mathcal{F},K) = \mathcal{F} + \Theta_{IFFT}({\rm Conv}(\sigma(K,\mathcal{F}) \odot \Theta_{FFT}(\mathcal{F})))
\label{equ:sa}
\end{equation}
\end{small}
where $\odot$ denotes low-pass filtering with adaptive Gaussian smoothed filters $\sigma(K,\mathcal{F})$, which has the same spatial size as $\mathcal{F}$, 
and $K$ is the bandwidth. 
To make $\sigma(K,\mathcal{F})$ input-aware, we create an initial 2D Gaussian map based on $K$, and apply pooling and linear layers on $\mathcal{F}$ to predict a scale parameter to update the Gaussian map.
Thanks to the spectral convolution theorem, the efficient point-wise spectral convolution globally updates $\mathcal{F}$. We treat the spectral-operated features as residuals and add them to the original input features for enhancement. Overall, SCF, which takes visual features $\mathcal{F}_{v}$ and word-level text features $\mathcal{F}_{w}$ as input, is computed as:
\begin{small}
\begin{equation}
{\rm SCF}(\mathcal{F}_{w},\mathcal{F}_{v}) = {\rm SA}({\rm SA}(\mathcal{F}_{v})\otimes {\rm Att}({\rm SA}(\mathcal{F}_{v}), \mathcal{F}_{w}))
\label{equ:SCF}
\end{equation}
\end{small}

\subsection{Conditional Patch Segmentation} 

We devise Conditional Patch Kernel (CPK) as the segmentation head to predict patch masks from the encoded vision-language features $\mathcal{F}_{vl}$ that are fully perceived by CPK. Unlike previous works~\cite{MTTR,ReferFormer}, CPK predicts a sequence of labels for each token rather than a single label, efficiently improving segmentation resolution along the channel dimension.

Specifically, we first use sentence-level text features $\mathcal{F}_{s}$ and multiple learnable embeddings to generate instance queries $Q \in{\mathbb{R} ^{N \times C}}$. Next, $Q$ is projected into instance embeddings $\mathcal{E} \in{\mathbb{R} ^{N \times C}}$ using the transformer decoder and $\mathcal{E}$ is leveraged to predict CPK for each instance query:
\begin{equation}
{\rm CPK}(Q,\mathcal{F}_{vl}) = \Theta({\rm FC}({\rm Att}(Q, \mathcal{F}_{vl})))
\label{equ:cpk}
\end{equation}
where $\Theta$ denotes the parameterization operation that reshapes CPK to form two point-wise convolutions with the output channel number of 16, which is similar to~\cite{ConditionConv}.
Since $Q$ changes dynamically according to different linguistic expressions, CPK becomes instance-aware and can separate objects of interest from $F_{vl}$.
Finally, we apply the parameterized CPK (dynamic point-wise convolutions) on $F_{vl}$ to predict patch masks $\mathcal{M}_{P} \in{\mathbb{R} ^{\frac{H}{i} \times \frac{W}{i} \times p^{2}}}$, where $\frac{H}{i} \times \frac{W}{i} $ denotes the spatial resolution of $F_{vl}$ and $p^2$ denotes the increased segmentation resolution on the channel dimension. 

During inference, we can reshape patch masks to $\mathcal{M}_{P} \in{\mathbb{R} ^{\frac{Hp}{i} \times \frac{Wp}{i}}}$ to efficiently generate fine-grained segmentation from low-resolution $F_{vl}$. The resolution of prediction will be consistent with the input when $p$ equals to $i$. 
We found that this efficient CPK can achieve competitive performance compared to methods that use heavy decoders.

\begin{figure}[t!]
\centering
\includegraphics[width=1\columnwidth]{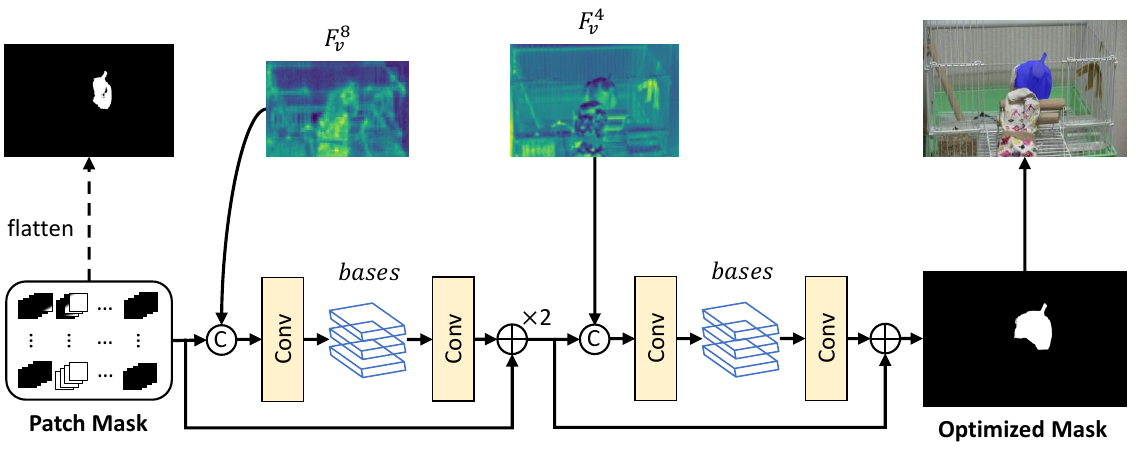}
\vspace{-4mm}
\caption{Multi-granularity Segmentation Optimizer, which predicts residual maps to optimize patch masks $\mathcal{M}_{P}$ progressively.
Flatten: reshape $\mathcal{M}_{P}$ from $\mathbb{R} ^{\frac{H}{i} \times \frac{W}{i} \times p^2}$ to $\mathbb{R} ^{\frac{Hp}{i} \times \frac{Wp}{i}}$ for visualization. $\times$2: upsampling operation.
}
\label{fig:refine}
\end{figure}

\subsection{Multi-granularity Segmentation Optimizer}
\label{subsec:MSO}

Segmenting encoded features $\mathcal{F}_{vl}$ with CPK avoids the detrimental drift effect on the segmentation head. However, visual details are required to produce accurate fine-grained masks. 
We propose Multi-granularity Segmentation Optimizer (MSO) to achieve this goal.

An overview of MSO is shown in Fig.~\ref{fig:refine}. It takes the predicted patch masks $\mathcal{M}_{P}$ as object priors and reuses visual features $\mathcal{F}_{v}$ with spatial strides of \{4,8\} to gradually recover visual details and refine the priors.
Specifically, MSO first concatenates $\mathcal{M}_{P}$ and $\mathcal{F}_{v}$ and projects them to low dimensional bases. 
Next, residual masks predicted by performing another convolution on these bases are used to correct $\mathcal{M}_{P}$.
Finally, the optimized patch masks achieve the input resolution by reshaping from $\mathbb{R} ^{\frac{H}{4} \times \frac{W}{4} \times 4^2}$ to $\mathbb{R} ^{H \times W}$.
Since MSO does not include heavy computations, the segment-and-optimize pipeline makes our approach perform better with efficient inference time.

\begin{table}[t!]
\setlength{\tabcolsep}{4.5pt}
\centering
\footnotesize
\begin{tabular}{lccc}
\toprule 
Methods & Single-frame & Multi-frames & Multi-objects \\ 
\midrule 
~\cite{URVOS,CMSA,PMINet} \emph{et al.} & \checkmark & & \\ 
~\cite{MTTR,ReferFormer,MLRL,YOFO} \emph{et al.} & \checkmark & \checkmark & \\
Fast \ourmethod{} (Ours) & \checkmark & \checkmark & \checkmark \\
\bottomrule 
\end{tabular}
\caption{Comparing different methods for their ability to segment single or multiple frames or multiple objects simultaneously.} \label{tab:multirvos} 
\vspace{-1mm}
\end{table}

\subsection{Multi-Object R-VOS}
Existing R-VOS methods perform single-frame (frame-wise) segmentation~\cite{URVOS,CMSA,PMINet} or multi-frame (clip-wise) segmentation~\cite{MTTR,ReferFormer,MLRL} for an {\em individual} referred object at a time. However, to the best of our knowledge, no existing work explores the simultaneous segmentation of {\em multiple} referred objects in video using a common GPU, which is important for real-world scenarios. To fill this gap, we present a new paradigm called multi-object R-VOS.

The key to multi-object R-VOS is designing a network that shares computationally intensive features for multiple objects, and enables different instance features to be decoupled before segmentation. 
To achieve this, we extend \ourmethod{} for multi-object R-VOS by introducing multi-instance fusion and decoupling. 
As shown in Table~\ref{tab:multirvos}, our method, dubbed Fast \ourmethod{}, can simultaneously segment multiple objects (in multiple frames) using a single 24GB GPU.

Fast \ourmethod{} shares visual features as well as vision-language features for all referred objects to make the network efficient, and decouples the shared features to make them instance-specific before the segmentation stage.
Firstly, visual features ($\mathcal{F}_{v}$) and language features ($\mathcal{F}_{w}$ and $\mathcal{F}_{s}$) are extracted. Next, we associate $\mathcal{F}_{v}$ and $\mathcal{F}_{w}$ using multi-instance fusion rather than the previous SCF. 
Multi-instance fusion is built on the foundation of SCF, which is depicted in Fig.~\ref{fig:fusion}. The difference is that multi-instance fusion
includes semantic fusion, which performs an element-wise add operation, after cross-attention to merge vision-language features of different expressions.
The features after semantic fusion perform Hadamard product with $\mathcal{F}_{v}$ to generate the vision-language features $\mathcal{F}_{vl}$ for all objects:
\vspace{-1mm}
\begin{small}
\begin{equation}
{\rm SF}(\mathcal{F}_{w},\mathcal{F}_{v})= \sum_{i=1}^{N} {\rm Att}(\mathcal{F}_{w}^{i},\mathcal{F}_{v})
\label{equ:mvlf1}
\end{equation}
\end{small}
\begin{small}
\begin{equation}
{\rm MIF}(\mathcal{F}_{w},\mathcal{F}_{v}) =  {\rm SA}({\rm SA}(\mathcal{F}_{v}) \otimes {\rm SF}(\mathcal{F}_{w},{\rm SA}(\mathcal{F}_{v})))
\label{equ:mvlf}
\end{equation}
\end{small}
where $\otimes$ denotes Hadamard product and $N$ denotes the number of expressions.
After vision-language fusion, we encode $\mathcal{F}_{vl}$ using the transformer encoder to enrich its semantic information, and plug multi-instance decoupling to decouple features for each instance. 
Multi-instance decoupling employs $\mathcal{F}_{w}$ and cross-attention to decouple $\mathcal{F}_{vl}$ to predict instance embeddings $\mathcal{E}$ for different referred objects. These embeddings are then projected to CPKs to predict the patch masks.
Thus, FAST \ourmethod{} shares features, which account for most of the computational overhead, for different expressions, making it efficient for referring segmentation.

\begin{table*}[t!]
\setlength{\tabcolsep}{8pt}
\centering
\small
\begin{tabular}{l | c | c | c c c c | c c c}
\toprule 
\multirow{2}{*}{Method} & \multirow{2}{*}{Year} & \multirow{2}{*}{Backbone} & \multicolumn{4}{c |}{Ref-YouTube-VOS} & \multicolumn{3}{c}{Ref-DAVIS17} \\
 & &  & \( \mathcal{J} \)\&\( \mathcal{F} \) & \( \mathcal{J} \) & \( \mathcal{F} \) & FPS &  \( \mathcal{J} \)\&\( \mathcal{F} \) & \( \mathcal{J} \) & \( \mathcal{F} \) \\
\midrule 
CMSA~\cite{CMSA} & 2019 & ResNet-50 & 36.4 & 34.8 & 38.1 & - & 40.2 & 36.9 & 43.5 \\
URVOS~\cite{URVOS} & 2020 & ResNet-50 & 47.2 & 45.3 & 49.2 & - & 51.5 & 47.3 & 56.0 \\
CMPC-V~\cite{CMPC} & 2021 & I3D & 47.5 & 45.6 & 49.3 & - & - & - & - \\
PMINet~\cite{PMINet} & 2021 & ResNeSt-101 & 53.0 & 51.5 & 54.5 & - & - & - & - \\
YOFO~\cite{YOFO} & 2022 & ResNet-50 & 48.6 & 47.5 & 49.7 & 10  & 53.3 & 48.8 & 57.8\\
LBDT~\cite{LBDT} & 2022 & ResNet-50 & 49.4 & 48.2 & 50.6 & -  & 54.3 & - & - \\
MLRL~\cite{MLRL} & 2022 & ResNet-50 & 49.7 & 48.4 & 51.0 & -  & 52.8 & 50.0 & 55.4\\ 
MTTR~\cite{MTTR} & 2022 & Video-Swin-T & 55.3 & 54.0 & 56.6 & - & - & - & -\\ 
MANet~\cite{MANet} & 2022 & Video-Swin-T & 55.6 & 54.8 & 56.5 & - & - & - & - \\

ReferFormer~\cite{ReferFormer} & 2022 & Video-Swin-T & 56.0 & 54.8 & 57.3 & 50 & - & - & - \\
\rowcolor[gray]{0.9} 
\ourmethod{} (Ours)  & 2023 & Video-Swin-T  & \textbf{58.9} & \textbf{57.7} & \textbf{60.0} & \textbf{65} & \textbf{56.7} & \textbf{53.3} & \textbf{60.0}\\

\midrule
\multicolumn{10}{l}{Pre-training with RefCOCO/+/g \& larger backbone} \\
\midrule

ReferFormer~\cite{ReferFormer} & 2022 & Video-Swin-T & 59.4 & 58.0 & 60.9 & 50 & 59.6 & 56.5 & 62.7  \\
\rowcolor[gray]{0.9} 
\ourmethod{} (Ours)  & 2023 & Video-Swin-T  & \textbf{62.0} & \textbf{60.4} & \textbf{63.5} & \textbf{65} & \textbf{61.9} & \textbf{59.0} & \textbf{64.8} \\ 

ReferFormer~\cite{ReferFormer} & 2022 & Video-Swin-B & 62.9 & 61.3 & 64.6 & 33 & 61.1 & 58.1 & 64.1 \\
\rowcolor[gray]{0.9} 
\ourmethod{} (Ours)  & 2023 & Video-Swin-B  & \textbf{65.7} & \textbf{63.9} & \textbf{67.4} & \textbf{40} & \textbf{63.3} & \textbf{60.6} & \textbf{66.0} \\ 
\bottomrule 
\end{tabular}
\caption{Quantitative comparison to state-of-the-art methods on the validation split of Ref-YouTube-VOS and Ref-DAVIS17.
} \label{tab:quantitative}
\end{table*}

\begin{table*}[t!]
\small
\centering
\begin{tabular}{l | c | c c c | c c c }
\toprule
\multirow{2}{*}{Method} & \multirow{2}{*}{Backbone} & \multicolumn{3}{c |}{A2D-Sentences} & \multicolumn{3}{c}{JHMDB-Sentences} \\
 & & mAP & Overall IoU & Mean IoU & mAP & Overall IoU & Mean IoU \\
\midrule
Hu \etal ~\cite{HuA2D} & VGG-16 & 13.2 & 47.4 & 35.0 & 17.8 & 54.6 & 52.8 \\
Gavrilyuk \etal ~\cite{GavrilyukA2D}  & I3D & 19.8 & 53.6 & 42.1 & 23.3 & 54.1 & 54.2 \\ 
ACAN~\cite{ACAN} & I3D & 27.4 & 60.1 & 49.0 & 28.9 & 57.6 & 58.4  \\
CMPC-V ~\cite{CMPC} & I3D & 40.4 & 65.3 & 57.3  & 34.2 & 61.6 & 61.7 \\
ClawCraneNet~\cite{Clawcranenet} & ResNet-50/101 & - & 63.1 & 59.9 & - & 64.4 & 65.6  \\
MTTR~\cite{MTTR} & Video-Swin-T & 46.1 & 72.0 & 64.0 & 39.2 & 70.1 & 69.8  \\
ReferFormer~\cite{ReferFormer} & Video-Swin-T & 52.8 & 77.6 & 69.6 & 42.2 & 71.9 & 71.0  \\ 
\rowcolor[gray]{0.9} 
\ourmethod{} (Ours) & Video-Swin-T & \textbf{56.1} & \textbf{78.0} & \textbf{70.4} & \textbf{44.4} & \textbf{72.8} & \textbf{71.7} \\ 

ReferFormer~\cite{ReferFormer} & Video-Swin-B & 55.0 & 78.6 & 70.3 & 43.7 & 73.0 & 71.8   \\ 
\rowcolor[gray]{0.9} 
\ourmethod{} (Ours) & Video-Swin-B & \textbf{58.5} & \textbf{79.9} & \textbf{72.0} & \textbf{45.0} & \textbf{73.7} & \textbf{72.5}  \\ 
\bottomrule
\end{tabular}
\caption{Quantitative comparison to state-of-the-art R-VOS methods on A2D-Sentences and JHMDB-Sentences.} \label{tab:quantitativea2d}
\end{table*}

\subsection{Instance Matching and Loss Functions}
\label{sec:matchloss}

Following~\cite{MTTR,ReferFormer}, we perform instance matching with $N=5$ learnable instance queries to improve fault tolerance. These queries are projected to CPKs to predict $N$ potential patch masks $\mathcal{M}_{P}$ for each expression. 
The Hungarian algorithm~\cite{Hungarian} is then adopted to select the best match based on the matching loss for training. During inference, we directly employ the predicted confidence scores $\mathcal{S}$ to measure the instance queries and select the results. 

We adopt the same training losses and weights as used in ~\cite{ReferFormer,DeformableTransformer} for a fair comparison. Specifically, we use Dice loss~\cite{DICE} and Focal loss~\cite{FOCAL} for patch mask $\mathcal{M}_{P}$ and optimized mask $\mathcal{M}_{O}$, Focal loss~\cite{FOCAL} for confidence scores $\mathcal{S}$, and L1 and GIoU~\cite{GIOU} loss for bounding boxes $\mathcal{B}$. The final training loss functions are:
\begin{small}
\begin{equation}
\mathcal{L}_{train} = \lambda_{\mathcal{M}_{P}}\mathcal{L}_{\mathcal{M}_{P}} +
\lambda_{\mathcal{M}_{O}}\mathcal{L}_{\mathcal{M}_{O}} +
\lambda_{\mathcal{B}}\mathcal{L}_{\mathcal{B}} +  \lambda_{\mathcal{S}}\mathcal{L}_{\mathcal{S}} 
\label{equ:losstrain}
\end{equation}
\end{small}
where $\mathcal{L}$ and $\lambda_{}$ are the loss term and weight, respectively.

\section{Experiments}

\subsection{Datasets and Metrics}
\noindent\textbf{Datasets.} 
We evaluate \ourmethod{} on four video benchmarks: Ref-YouTube-VOS~\cite{URVOS}, Ref-DAVIS17~\cite{RefDAVIS}, A2D-Sentences~\cite{GavrilyukA2D}, and JHMDB-Sentences~\cite{JHMDB}. Ref-YouTube-VOS is currently the largest dataset for R-VOS, containing 3,978 videos with about 13K expressions. 
Ref-DAVIS17 is an extension of DAVIS17~\cite{DAVIS} by including the language expressions of different objects and contains 90 videos. 
A2D-Sentences is a general actor and action segmentation dataset with over 3.7K videos and 6.6K action descriptions. JHMDB-Sentences includes 928 videos and 928 descriptions covering 21 different action classes.

\noindent\textbf{Evaluation Metrics.} 
We adopt the standard metrics to evaluate our models: region similarity $\mathcal{J}$ (average IoU), contour accuracy $\mathcal{F}$ (average boundary similarity), and their mean value \( \mathcal{J} \)\&\( \mathcal{F} \). All results are evaluated using the official code or server. 
On A2D-Sentences and JHMDB-Sentences, we adopt mAP, overall IoU, and mean IoU for evaluation.

\subsection{Implementation Details}

Following ~\cite{MTTR,MANet,ReferFormer}, we train our models on the training set of Ref-YouTube-VOS, and directly evaluate them on the validation split of Ref-YouTube-VOS and Ref-DAVIS17 without any additional techniques, \eg, model ensemble, joint training, and mask propagation, since they are not the focus of this paper. 
Additionally, we present results for our models first pre-trained on RefCOCO/+/g~\cite{RefCOCO2,RefCOCO} and then fine-tuned on Ref-YouTube-VOS.
Similar to~\cite{ReferFormer,DeformableTransformer}, we set the coefficients for different losses $\lambda_{dice}$, $\lambda_{focal}$, $\lambda_{L1}$, $\lambda_{giou}$ to 5, 2, 5, 2, respectively. The models are trained using 2 RTX 3090 GPUs with 5 frames per clip for 9 epochs. All frames are resized to have the longest side of 640 pixels. Further implementation details are in the supplementary material.

\subsection{Quantitative Results}
\label{sec:QuantitativeResults}
\noindent\textbf{Ref-YouTube-VOS and Ref-DAVIS17.} 
We compare \ourmethod{} with recently published works in Table~\ref{tab:quantitative}. Our approach surpasses present solutions on the two datasets across all metrics.
\textcolor{black}{On Ref-YouTube-VOS, \ourmethod{} with the Video Swin Tiny backbone achieves 58.9 \( \mathcal{J} \)\&\( \mathcal{F} \) at 65 FPS, which is \textbf{2.9\%} higher and \textbf{1.3$\times$} faster than the previous state-of-the-art ReferFormer~\cite{ReferFormer}. 
Our approach runs faster due to the use of the segment-and-optimize pipeline, which avoids the need for heavy feature decoders.
When pre-training with RefCOCO/+/g and using a larger backbone, \ie, Video Swin Base, the performance of \ourmethod{} further boosts to 65.7 \( \mathcal{J} \)\&\( \mathcal{F} \), consistently leading all other solutions by more than \textbf{2.8\%}.
On Ref-DAVIS17, \ourmethod{} achieves 63.3 \( \mathcal{J} \)\&\( \mathcal{F} \), outperforming state-of-the-art by \textbf{2.2\%} and demonstrating the generality of our approach.}

\noindent\textbf{A2D-Sentences and JHMDB-Sentences.} 
\textcolor{black}{We further evaluate \ourmethod{} on A2D-Sentences and JHMDB-Sentences in Table~\ref{tab:quantitativea2d}. Following~\cite{ReferFormer}, the models are first pre-trained on RefCOCO/+/g and then fine-tuned on A2D-Sentences. JHMDB-Sentences is used only for evaluation. As shown in Table~\ref{tab:quantitativea2d}, \ourmethod{} achieves superior performance compared to other state-of-the-art R-VOS methods and surpasses the nearest competitor Referformer~\cite{ReferFormer} by \textbf{3.5/1.3\%} mAP on A2D-Sentences and JHMDB-Sentences, respectively.}

\begin{table}[t!]
\setlength{\tabcolsep}{4.5pt}
\centering
\footnotesize
\begin{tabular}{l c c c c c c c}
\toprule 
Method & \multicolumn{3}{c}{Ref-DAVIS17} & \multicolumn{4}{c}{Ref-YouTube-VOS} \\
\midrule 
 & \( \mathcal{J} \)\&\( \mathcal{F} \) & \( \mathcal{J} \) & \( \mathcal{F} \) & \( \mathcal{J} \)\&\( \mathcal{F} \) & \( \mathcal{J} \) & \( \mathcal{F} \) & FPS \\
\midrule 
ReferFormer~\cite{ReferFormer} & 54.5 & 51.0 & 58.0 & 56.0 & 54.8 & 57.3 & 50  \\  
Fast \ourmethod{} (Ours) & 54.2 & 51.1 & 57.3 & 54.2  & 53.1  & 55.3  & \textbf{185} \\
\bottomrule 
\end{tabular}
\caption{Evaluation of Fast \ourmethod{} on Ref-DAVIS17 and Ref-YouTube-VOS. Video-Swin-T is adopted as the backbone.} 
\label{tab:fast\ourmethod{}}
\end{table}

\begin{table}[t!]
\centering
\small
\begin{tabular}{c c c c c c c}
\toprule 
\multicolumn{3}{c}{Components} & \multicolumn{4}{c}{Performance} \\
\midrule 
CPK & MSO & SCF & \( \mathcal{J} \)\&\( \mathcal{F} \) &  \( \mathcal{J} \) & \( \mathcal{F} \) & FPS \\
\midrule 
 & & & 54.4 & 52.7 & 56.2 & 70 \\ 
\checkmark  & & & 55.8 & 54.5 & 57.1 & 70 \\
  & \checkmark & & 57.7 & 56.3 & 59.1 & 69 \\
 & & \checkmark & 55.8 & 54.3 & 57.4 & 66 \\ 
\checkmark & \checkmark & & 57.9 & 56.7 & 59.1 & 69 \\
\checkmark & \checkmark & \checkmark & 58.9 & 57.7 & 60.0 & 65 \\
\bottomrule 
\end{tabular}
\caption{Ablation of different components on Ref-YouTube-VOS.} \label{tab:abmodule} 
\end{table}

\noindent\textbf{Multi-object R-VOS.} 
We extend \ourmethod{} to perform multi-object R-VOS, which is more practical and efficient for deployment. 
Fast \ourmethod{} is trained on Ref-YouTube-VOS without pre-training or postprocessing techniques.
We benchmark Fast \ourmethod{} on Ref-YouTube-VOS and Ref-DAVIS17 using the commonly used Video Swin Tiny, and compare the results with the state-of-the-art R-VOS method, which performs single-object segmentation.

As shown in Table~\ref{tab:fast\ourmethod{}}, Fast \ourmethod{} achieves reasonable performance and runs about \textbf{3.7$\times$} faster (\textbf{185} \vs{} 50 FPS) compared to ReferFormer~\cite{ReferFormer}. It should be noted that each object in the above datasets contains multiple expressions. 
On Ref-DAVIS17, we group expressions to have only one expression per object within each group and segment all expressions in each group simultaneously since the object identity is given.
On Ref-YouTube-VOS, all expressions in a video are segmented simultaneously due to the lack of object identity, making it more challenging.

\subsection{Ablation Study for Different Components}
\label{sec:ablation}
We conduct ablation experiments to evaluate the effectiveness of different components in \ourmethod{}. The components are added to the baseline model step-by-step.

\noindent\textbf{Conditional Patch Kernel.} 
As shown in Table~\ref{tab:abmodule}, CPK boosts the performance by 1.4\% compared with the recent instance-aware conditional kernels~\cite{ReferFormer}. The sequential labels of each token predicted by CPK contain more fine-grained information, making the prediction more accurate.

\noindent\textbf{Multi-granularity Segmentation Optimizer.} We devise MSO to optimize the predicted patch masks. 
As shown in Table~\ref{tab:abmodule}, MSO improves the performance by 3.3\%, indicating the importance of fine-grained visual details in R-VOS.

\noindent\textbf{Spectrum-guided Cross-modal Fusion.} We present SCF to perform global interactions by operating in the spectral domain. In Table~\ref{tab:abmodule}, using SCF to replace the traditional cross-attention in~\cite{ReferFormer,Attention} improves the \( \mathcal{J} \)\&\( \mathcal{F} \) by 1.4\%. 
We consider SCF extracts important low-frequency features and facilitates multimodal understanding globally, which is suitable for R-VOS since locating referred objects requires understanding the global context and token relations.

\begin{table}[t!]
\centering
\footnotesize
\setlength{\tabcolsep}{2.8pt}
\begin{tabular}{l c c c c}
\toprule 
Settings & Drift & Pipeline & \( \mathcal{J} \)\&\( \mathcal{F} \) & FPS \\
\midrule 
Baseline + Decoder & \checkmark & decode-and-segment & 56.0 & 50 \\
Baseline + Decoder + MSO & \checkmark & decode-and-segment & 56.4 & 49 \\
Baseline + MSO & $\times$ & segment-and-optimize & \textbf{57.7} & \textbf{69} \\
\bottomrule 
\end{tabular}
\caption{Feature drift analysis using ReferFormer~\cite{ReferFormer} (Baseline + Decoder) and \ourmethod{} w/o CPK \& SCF (Baseline + MSO). 
Significant improvement is achieved by addressing the drift issue (last row). Adding MSO on top of ReferFormer to recover visual details (for a second time) still performs worse than our basic pipeline.
}
\label{tab:abdrift} 
\end{table}

\subsection{Ablation Study for Feature Drift}
We conduct ablation study in Table \ref{tab:abdrift} to demonstrate the feature drift problem.
Our segment-and-optimize pipeline addresses the adverse drift effect discussed in Section~\ref{sec:driftanalysis} to significantly outperform ReferFormer~\cite{ReferFormer} by 1.7\% points and runs 1.4$\times$ faster. 
Furthermore, adding MSO on top of ReferFormer still performs worse due to the negative drift impact caused by the decode-and-segment pipeline.
These results demonstrate the efficacy of our proposed segment-and-optimize pipeline.

\subsection{Referring Image Segmentation Results}
\label{sec:ris}
We apply \ourmethod{} to referring image (expression) segmentation without any architectural modifications, and compare against the current state-of-the-art methods on RefCOCO/+/g~\cite{RefCOCO2,RefCOCO}. 
A single \ourmethod{} model is trained on RefCOCO/+/g without large-scale pre-training.
As shown in Table~\ref{tab:ris}, \ourmethod{} achieves advanced performance on all three benchmarks. These results demonstrate the efficacy of \ourmethod{} in referring image segmentation.

\begin{table}[t]
\small
\centering
\begin{tabular}{l | c | c | c}
\toprule
Method &  RefCOCO & RefCOCO+ & RefCOCOg \\
\midrule
MaIL~\cite{MaIL} & 70.1 & 62.2 & 62.5 \\
CRIS~\cite{CRIS} & 70.5 & 62.3 & 59.9 \\
RefTR~\cite{RefTR} & 70.6 & - & - \\
LAVT~\cite{LAVT} & 72.7 & 62.1 & 61.2 \\
VLT~\cite{VLT} & 73.0 & 63.5 & 63.5 \\
\ourmethod{} (Ours) & \textcolor{black}{\textbf{76.3}} & \textbf{66.4} & \textbf{70.0} \\ 
\bottomrule
\end{tabular}
\caption{Quantitative evaluation on the validation split of RefCOCO/+/g. Overall IoU is adopted as the evaluation metric.} 
\label{tab:ris}
\end{table}

\begin{figure*}[t!]
\centering
\includegraphics[width=2\columnwidth]{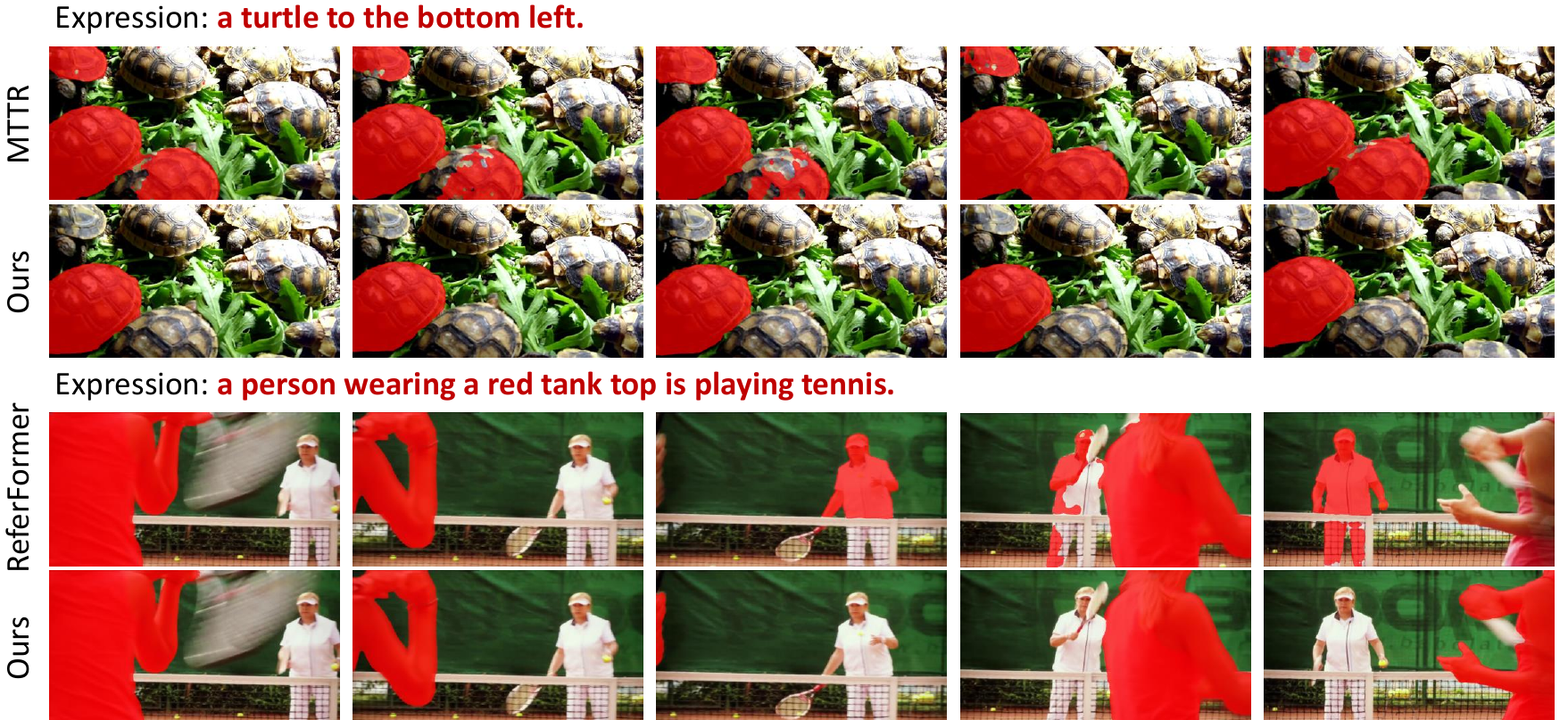}
\caption{Qualitative comparison of our method with others.}
\label{fig:qualitative}
\vspace{-2mm}
\end{figure*}

\subsection{Inference Time Analysis of Multi-Object RVOS}
\begin{figure}[t]
\centering
\includegraphics[width=1\columnwidth]{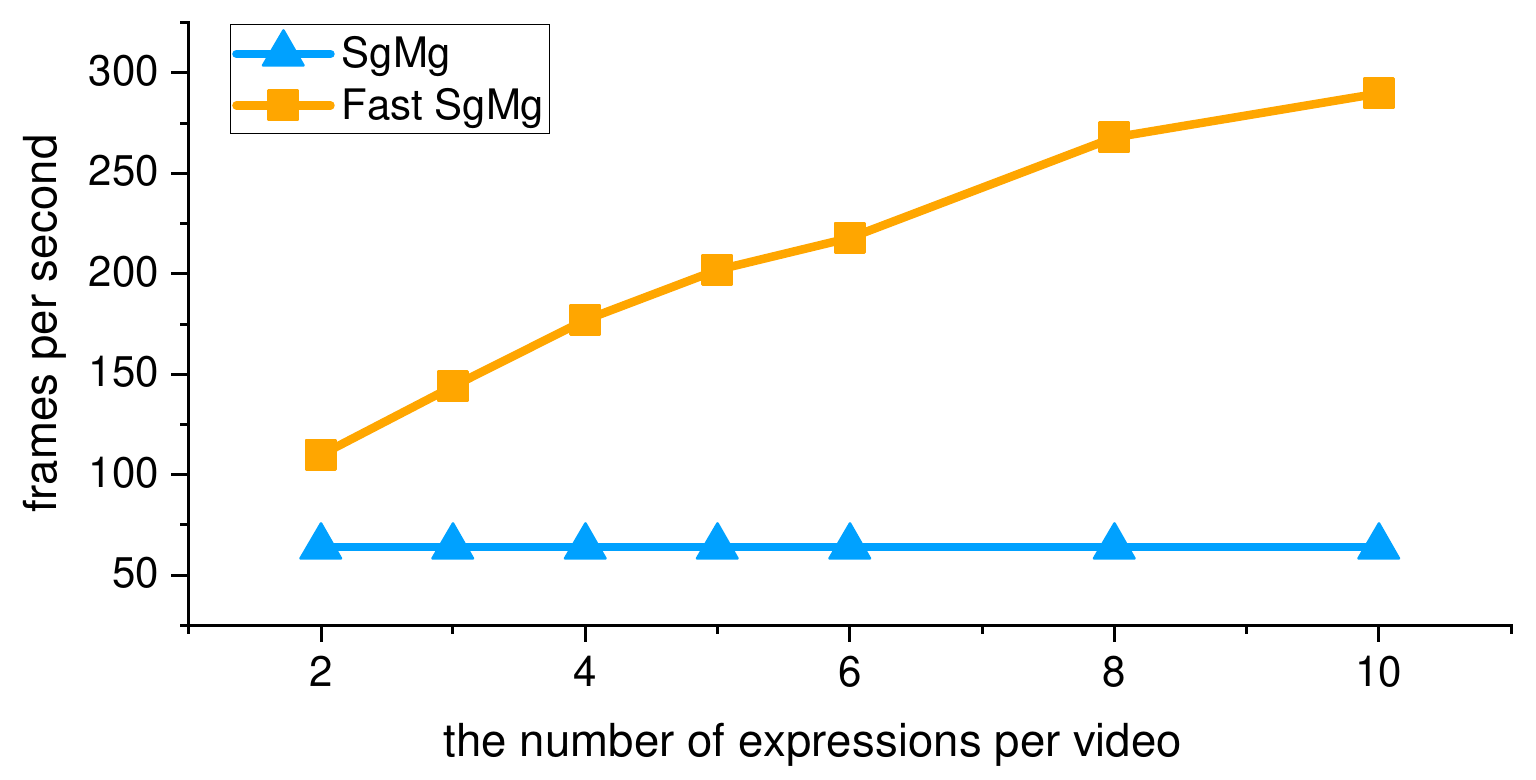}
\caption{Efficiency analysis of \ourmethod{} and Fast \ourmethod{} for videos with different numbers of expressions on Ref-YouTube-VOS.} \label{fig:fpsanalyze}
\end{figure}

We analyze the efficiency of the proposed multi-object R-VOS paradigm by comparing the FPS of Fast \ourmethod{} and \ourmethod{} on videos with different numbers of expressions.
As illustrated in Fig.~\ref{fig:fpsanalyze}, 
Fast \ourmethod{} performs about 2$\times$ faster than \ourmethod{} when there are two expressions per video on average. 
As the number of expressions increases, Fast \ourmethod{} achieves faster reasoning time per object per frame due to its utilization of the multi-object R-VOS paradigm. 
When there are ten expressions in each video, Fast \ourmethod{} performs at nearly 300 FPS, which is about \textbf{5$\times$} faster than \ourmethod{}.

\subsection{Qualitative Results}

In Fig.~\ref{fig:qualitative}, we show qualitative comparison with ReferFormer~\cite{ReferFormer} and MTTR~\cite{MTTR}. \ourmethod{} can handle different objects of the same category or with the same behavior.

\subsection{Feature Visualization of SCF}

\begin{figure}[t]
\centering
\includegraphics[width=1\columnwidth]{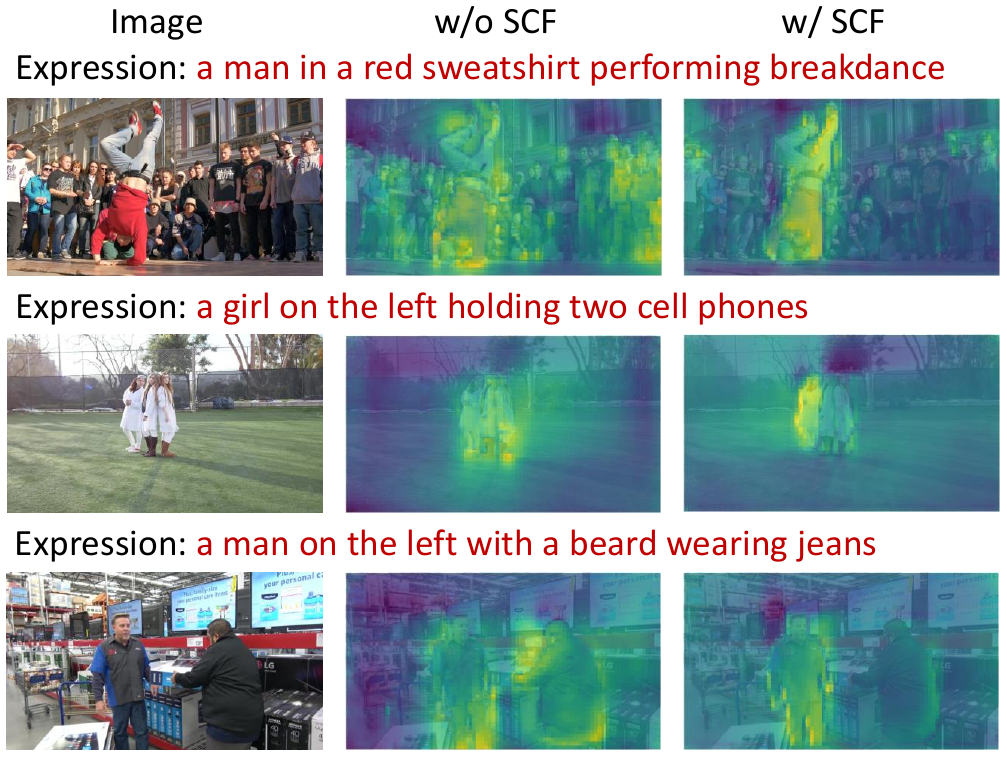}
\caption{Visualization of the vision-language features extracted w/o and w/ our SCF.}
\label{fig:fusion_visualize}
\vspace{-1mm}
\end{figure}

In Fig.~\ref{fig:fusion_visualize}, we visualize the vision-language features extracted by our SCF in comparison to the cross-attention used in~\cite{ReferFormer}. The features extracted by SCF exhibit superior grounding ability in locating target objects, resulting in better performance for \ourmethod{}.

\section{Conclusion}
We discovered the feature drift issue in current referring video object segmentation (R-VOS) methods, which negatively affects the segmentation kernels. We presented \ourmethod{}, a novel segment-and-optimize approach for R-VOS that avoids the drift issue and optimizes masks with visual details. We also provided a new perspective to encourage vision-language global interactions in the spectral domain with Spectrum-guided Cross-modal Fusion. 
Additionally, we proposed the multi-object R-VOS paradigm by extending \ourmethod{} with multi-instance fusion and decoupling. 
Finally, we evaluated our models on four video benchmarks and demonstrated that our approach achieves state-of-the-art performance on all four datasets.

\vspace{2mm}
\noindent\textbf{Acknowledgment.}
This research was supported by the Australian Research Council Industrial Transformation Research Hub IH180100002. Professor Ajmal Mian is the recipient of an Australian Research Council Future Fellowship Award (project number FT210100268) funded by the Australian Government.


\clearpage
{\small
\bibliographystyle{ieee_fullname}
\bibliography{egbib}
}


\clearpage
\appendix

\renewcommand\thesection{\Alph{section}}
\renewcommand\thefigure{\Alph{figure}}
\renewcommand\thetable{\Alph{table}}

\section{Encoder and Transformer Details}

\noindent\textbf{Visual Encoder.}
Video Swin Transformer~\cite{VideoSwin} is adopted as the visual encoder because of its effectiveness in extracting robust spatio-temporal features. Multi-stage visual features with spatial strides of \{4,8,16,32\} are used for segmentation, \ie, the last three stages for cross-modal fusion and the first two stages for multi-granularity optimization. 
We resize the multi-scale vision-language features to the same resolution and use element-wise addition to integrate them into a single layer for conditional segmentation.

\noindent\textbf{Textual Encoder.}
The pre-trained RoBERTa~\cite{Roberta} is used to encode language expressions due to its proven performance in natural language processing tasks. Each expression is encoded into word features and sentence features.

\noindent\textbf{Transformer.}
Deformable Transformer~\cite{DeformableTransformer} with 4 encoder and decoder layers is used to encode vision-language features and predict instance embeddings due to its effectiveness and efficiency in capturing global pixel-level relations.

\section{Instance Matching Details}
Our instance matching process follows the standard paradigm used by previous transformer-based methods for video segmentation~\cite{ReferFormer,MTTR,mask2former,vita,VISTR,idol}. Specifically, we use $N$=5 learnable instance queries for prediction and apply the Hungarian algorithm~\cite{Hungarian} to select the best result.
To achieve this, \ourmethod{} predicts patch masks $\mathcal{M}_{P}$, bounding boxes $\mathcal{B}$, and confidence scores $\mathcal{S}$ for each expression. Using the set of predictions $y = \{\forall y^{i}, i \in [1, ..., N] \}$, where $y^{i} = {\{\mathcal{M}_{P}^{i,j}, \mathcal{B}^{i,j}, \mathcal{S}^{i,j}\}}_{j=1}^{T}$, we compute the matching loss $\mathcal{L}_{match}$ for each query based on the ground truth $\hat{y}$ and employ Hungarian algorithm to find the best match that has the minimum loss. $\mathcal{L}_{match}$ lies in three parts:
\begin{equation}
\mathcal{L}_{match} = \lambda_{\mathcal{M}_{P}}\mathcal{L}_{\mathcal{M}_{P}} + \lambda_{\mathcal{B}}\mathcal{L}_{\mathcal{B}} +  \lambda_{\mathcal{S}}\mathcal{L}_{\mathcal{S}} 
\label{equ:lossmatch}
\end{equation}
where $\lambda$ denotes the coefficient to balance $\mathcal{L}_{match}$.

\section{Further Implementation Details}
Our training settings follow~\cite{ReferFormer,YOFO,MTTR}.
The data augmentation includes random resize, random crop, random horizontal flip, and photometric distortion.
The models are trained using AdamW~\cite{AdamW} optimizer for 12 epochs during pre-training, and 6 or 9 epochs during main training depending on whether pre-training is used.
During pre-training on RefCOCOs, we set the initial learning rates of 2.5e-6, 1.25e-5, and 2.5e-5 for the text encoder, visual encoder, and the rest of the model, respectively. The pre-training employs a single frame, with the learning rates decayed by a factor of 10 at the 8th and 10th epochs.
In the main training, we freeze the text encoder, and the initial learning rates of 2.5e-5 and 5e-5 are adopted for the visual encoder and the rest, respectively. The learning rates are divided by 10 at the 6th and 8th epoch.

During inference, we perform clip-wise segmentation as in~\cite{ReferFormer}. Specifically, we set the clip length equal to the number of video frames for Ref-YoutubeVOS and 36 for Ref-DAVIS17 to enable better spatio-temporal feature representation and efficiency. 
Notably, our approach can also perform frame-wise segmentation to achieve good performance according to the referring image segmentation results presented in the main paper.

\section{Conditional Patch Segmentation}

\begin{figure}[h]
\centering
\includegraphics[width=1\columnwidth]{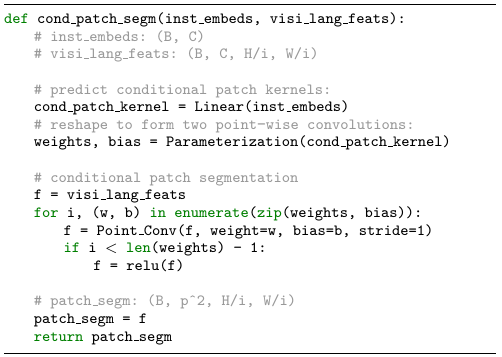}
\caption{Pseudo-code of conditional patch segmentation.}
\label{fig:cond_conv}
\end{figure}

We present the pseudo-code of our conditional patch segmentation process in Fig.~\ref{fig:cond_conv}.
To be specific, instance embeddings are employed to predict conditional patch kernels. The conditional patch kernels are reshaped to dynamic weights and bias, which form two point-wise convolutions. Finally, point-wise convolutions are used to segment vision-language features to obtain patch masks.

\section{Ablation of Spectral Convolutions in SCF}

We replace the spectral convolutions in Spectrum-guided Cross-modal Fusion (SCF) with spatial convolutions or linear layers, which contain more parameters than ours. As shown in Table~\ref{tab:fps}, our SCF that operates in the spectral domain achieves the best performance.

\begin{table}[h]
\small
\centering
\begin{tabular}{l | c | c }
\toprule
Module & \( \mathcal{J} \)\&\( \mathcal{F} \) & Para. Num. (M) \\
\midrule
SCF w/ Spatial Conv & 57.6  & 4.7 \\
SCF w/ Linear & 57.9  & 2.4 \\
SCF (Ours) & \textbf{58.9} & \textbf{2.4} \\
\bottomrule
\end{tabular}
\vspace{1mm}
\caption{Ablation of SCF with different operations.} 
\label{tab:fps}
\end{table}

\section{Additional t-SNE Visualizations}
\begin{figure}[h]
\centering
\includegraphics[width=0.9\columnwidth]{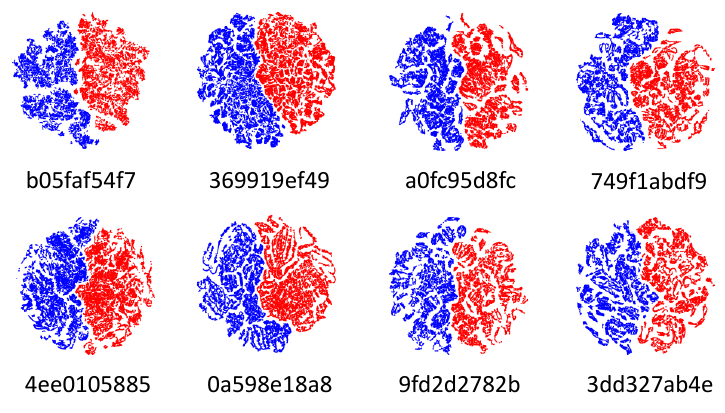}
\caption{t-SNE~\cite{tsne} visualization of the feature embeddings in different videos before (red cluster) and after (blue cluster) decoding.}
\label{fig:drift}
\end{figure}

To further demonstrate the presence of feature drift, we present additional t-SNE~\cite{tsne} visualizations in Fig.~\ref{fig:drift}. Specifically, we add the feature decoding process into the model, where the token embeddings of encoded features $\mathcal{F}_{vl}$ are decoded using the decoder in~\cite{ReferFormer} to obtain $\mathcal{F}_{vl}^{d}$ for all frames in each video. 
By visualizing these embeddings with t-SNE, we observe that the token embeddings of $\mathcal{F}_{vl}$ and $\mathcal{F}_{vl}^{d}$ are separated into two distinct clusters. 
This indicates that the decoding process results in feature drift.
However, the segmentation kernels struggle to perceive this drift during forward propagation since the kernels are predicted before the feature decoding.

\section{Additional Qualitative Results}

\begin{figure}[h!]
\centering
\includegraphics[width=1\columnwidth]{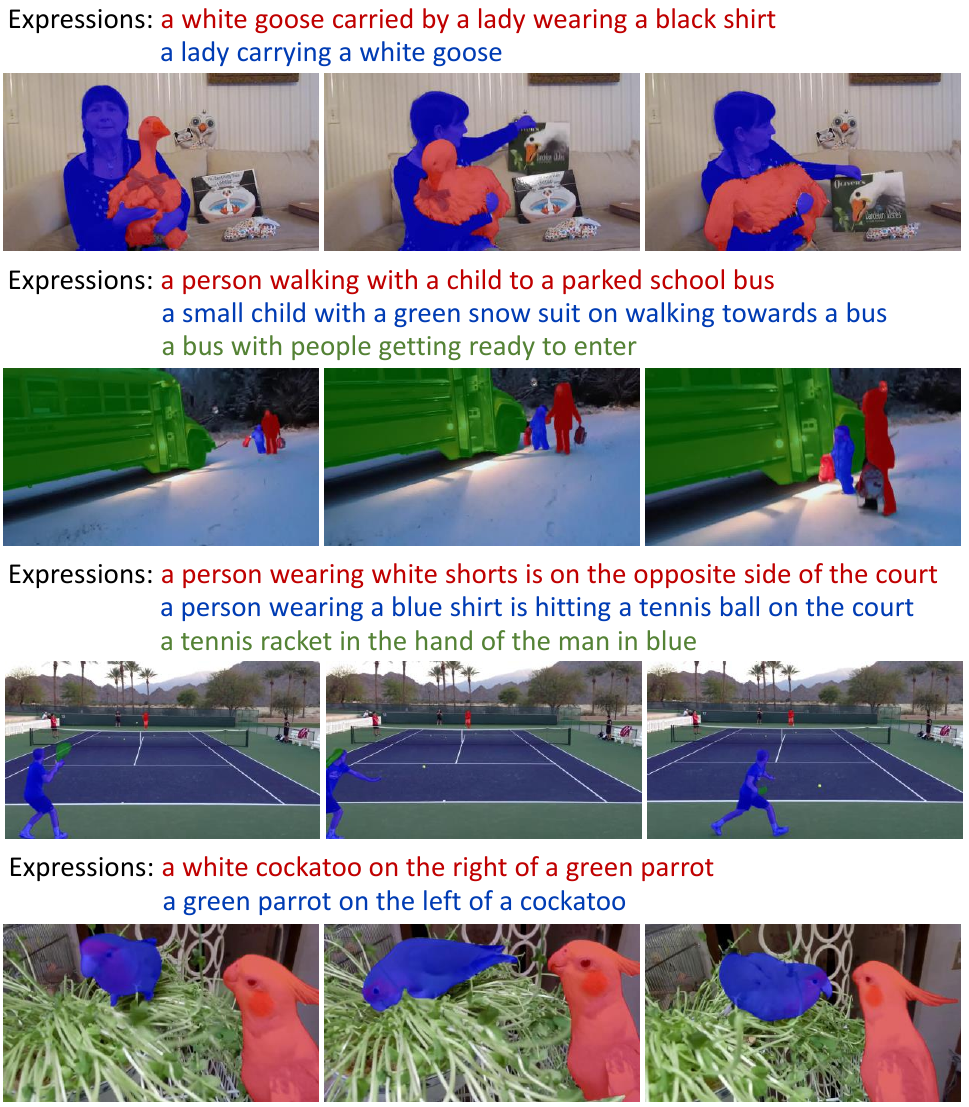}
\caption{Additional qualitative results of \ourmethod{}.}
\label{fig:SuppQualitative}
\end{figure}

In Fig.~\ref{fig:SuppQualitative}, we present additional qualitative results that include occlusion, similar appearance, fast motion, and small objects.

\end{document}